\documentclass[runningheads]{llncs}
\usepackage[T1]{fontenc}
\usepackage{graphicx}
\usepackage{enumitem}
\usepackage{amsmath}
\usepackage{amssymb}
\usepackage{multirow}

\usepackage[T1]{fontenc}
\usepackage{graphicx}
\usepackage[
linesnumbered,ruled,vlined]{algorithm2e}
\usepackage{algpseudocode}
\usepackage{algorithmicx}
\usepackage{algcompatible}
\usepackage{bbm}
\usepackage{amsmath}
\usepackage{textcomp}
\usepackage{amssymb}
\usepackage{authblk}
\usepackage{caption}
\usepackage{xcolor}
\usepackage{subcaption}

\usepackage[skip=2pt]{caption}

\usepackage{wrapfig}
\usepackage{tikz}
\usepackage{booktabs}

\usepackage{multirow}

\newcommand\blfootnote[1]{%
  \begingroup
  \renewcommand\thefootnote{}\footnote{#1}%
  \addtocounter{footnote}{-1}%
  \endgroup
}
\usepackage[pagebackref,breaklinks,colorlinks]{hyperref}
\usepackage{cleveref}
\usepackage{color}

\begin{document}
%
%\title{Consistent 3D Scene Editing via Semantic-Geometric Feature Fields in Gaussian Splatting}
%\title{Consistent Instance Segmentation in 3D Gaussian Splatting with Semantic–Geometric Guidance}
\title{Consistent Scene Understanding in 3D Gaussian Splatting via Multi-Cue Mask Refinement}
\titlerunning{Multi-Cue Mask Refinement in 3DGS}
\author{Hyunjoon Park\and
Donghyeon Cho*}
\authorrunning{Hyunjoon Park and Donghyeon Cho}
\institute{Department of Computer Science, Hanyang University, Seoul, South Korea \\
\email{\{junippini83, doncho\}@hanyang.ac.kr}}
\maketitle
\blfootnote{* Corresponding author. 
\\ Project page: \url{https://hjpark83.github.io/consisGS.github.io}. 
\\ Code: \url{https://github.com/hjpark83/Consistent-Scene-Understanding-in-3DGS-via-MCM}.
}
\begin{abstract}
Reliable instance-level scene understanding is a fundamental prerequisite for object-level interactions and high-fidelity 3D representations.
%
% While current methods often leverage 2D foundation models, such as the Segment Anything Model (SAM), to obtain these priors, 2D centric design of SAM typically yields fragmented masks and inconsistent predictions across different views.
While current methods often leverage 2D foundation segmentation models to obtain these priors, their 2D-centric design typically yields fragmented masks and inconsistent predictions across different views.
%
% To address these issues, we propose a multi-cue mask refinement framework that produces consistent 2D instance masks to guide the optimization of 3D Gaussian Splatting (3DGS) feature fields.
To address these issues, we propose a novel framework that produces consistent 2D instance masks to guide the optimization of 3D Gaussian Splatting (3DGS) feature fields.
Our framework consists of three main stages. 
(1) Multi-Cue Extraction that generates synergistic semantic, geometric, and structural priors from input images.
(2) Multi-Cue-Guided Mask Merging process that consolidates fragmented masks using a composite merge score derived from semantic, depth, and edge cues. 
(3) Cross-View Mask Matching that establishes globally consistent identity assignments across all viewpoints.
By transforming viewpoint-specific segments into coherent 3D primitives, our approach enables stable 3D instance segmentation and effective downstream editing tasks.
Experiments demonstrate that our method significantly improves cross-view consistency and segmentation stability over existing baselines while maintaining high-fidelity photometric reconstruction.
\end{abstract}
\begin{figure}[!t]
    \centering
    \includegraphics[width=1\linewidth]{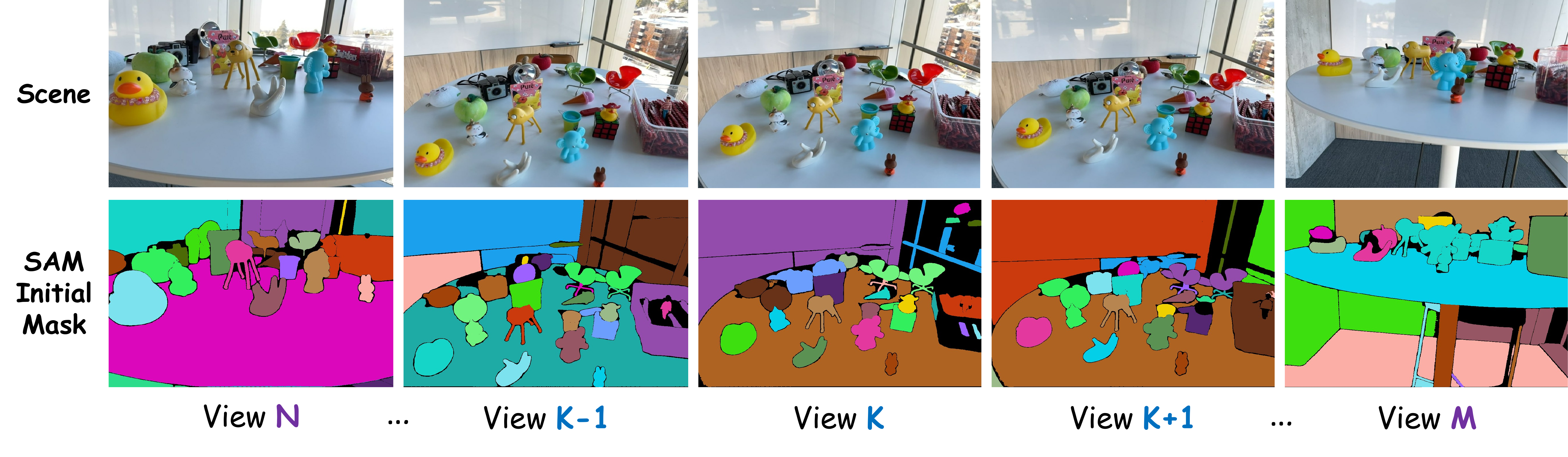}
    \caption{
    \textbf{Cross-view inconsistency from SAM over-segmentation.} 
    Automatic masks from SAM~\cite{SAM} suffer from inconsistent identities across distant ($N \dots M$) and even contiguous frames ($K \pm 1$). 
    Its 2D-centric nature induces over-segmentation and erroneous merging of distinct objects, impeding coherent 3D scene understanding.
    }
    \label{fig:problem1}
\end{figure}

\section{Introduction}
\label{sec:intro}

While Neural Radiance Fields (NeRF)~\cite{NeRF} pioneer high-fidelity scene representation through continuous volumetric functions, 3D Gaussian Splatting (3DGS)~\cite{3DGS} shifts the paradigm toward explicit primitive-based representations. 
Unlike implicit weight-based encoding of NeRF, 3DGS utilizes a discrete collection of Gaussians as a structured substrate, facilitating seamless integration with 2D foundation models like Segment Anything Model (SAM)~\cite{SAM}, GroundingDINO~\cite{groundingdino} and Contrastive Language-Image Pretraining (CLIP)~\cite{CLIP}. 
However, integrating these models into 3DGS remains challenging because 2D models lack 3D spatial awareness and geometric context and often produce inconsistent semantics across views, making the transition from fragmented 2D segmentations to coherent 3D primitives unstable.
Among these models, SAM is a representative 2D segmentation prior that is widely adopted for 3DGS-based scene editing. 
%
% The SAM offers dense, class-agnostic masks with precise boundaries for identifying editable object instances, which directly affects the 3D reasoning quality.
%
It offers dense, class-agnostic masks with precise boundaries for identifying editable object instances, which directly affects the 3D reasoning quality.
Despite these advantages, the automatic mask generation of SAM often suffers from severe over-segmentation, partitioning a single object into numerous disconnected fragments (Fig.~\ref{fig:problem1}).
%
% This fragmentation stems from a 2D-centric design of SAM, which lacks depth awareness and cross-view semantic consistency.
%
This fragmentation stems from its 2D-centric design, which lacks depth awareness and cross-view semantic consistency.
Consequently, visually similar but disjoint regions are erroneously merged, while minor appearance variations trigger unnecessary splits. 
%
% Such inaccuracies hinder the establishment of stable object identities and lead to cross-view mask ID inconsistencies.
%
Such inaccuracies hinder the establishment of stable object identities and lead to global ID inconsistencies.
To address these challenges, we reinterpret the SAM output as an initially overcomplete decomposition that is subsequently refined using multiple cues. 
%
% Specifically, we leverage three complementary cues: (1) DINOv2~\cite{dinov2} embeddings, which offer stronger object-level semantics than SAM; (2) a depth map estimated by a monocular model~\cite{Depthv2}, which provides geometric context that is absent from purely appearance-based segmentation; and (3) LoG-filtered edge maps, which localize salient structural boundaries.
%
Specifically, we leverage three complementary cues: (1) DINOv2~\cite{dinov2} embeddings, which offer strong object-level semantics; (2) Monocular depth maps~\cite{Depthv2}, which provides geometric context that is absent from purely appearance-based segmentation; and (3) LoG-filtered edge maps, which localize salient structural boundaries.
Based on these cues, our model performs Multi-Cue-Guided Mask Merging (MCM) to mitigate the over-segmentation produced by SAM. 
In this process, depth acts as a hard constraint to prevent erroneous merges across disjoint boundaries, while semantic and structural affinities unify texture-induced splits. 
%
% This resolves the "\textit{white-on-white dilemma}", preserving distinct surfaces at varying depths while consolidating fragmented objects.
%
This resolves the "\textit{white-on-white dilemma}", preserving distinct surfaces at varying depths while merging fragments.
Subsequently, globally consistent identities established via cross-view mask matching are lifted into 3D using a robust multiview consensus procedure utilizing majority voting.
Finally, joint optimization balances photometric fidelity with semantic regularization, yielding structured 3D masks as "\textit{actionable primitives}" for high-fidelity scene understanding.
Our contributions are as follows:

\begin{itemize}[itemsep=0pt, leftmargin=*]
    \item Multi-Cue Mask Refinement framework that mitigates SAM-induced over-segmentation based on multiple cues.
    % \item A cross-view mask matching method that establishes globally consistent object identities and suppresses view-dependent label inconsistencies across all viewpoints.
    \item Cross-view mask matching that establishes globally consistent object IDs and suppresses view-dependent label inconsistencies across all viewpoints.
    % \item A multiview consensus lifting procedure that transfers 2D identities to 3D Gaussians while filtering unreliable assignments through majority voting and variance analysis.
    \item Feature lifting that transfers 2D identities to 3D Gaussians while filtering unreliable assignments through majority voting and variance analysis.
    \item Extensive evaluation on LERF~\cite{LERF}, Replica~\cite{straub2019replica} and several real-world scenes, demonstrating substantial gains in correct object assignment and a significant reduction in over-segmented object counts.
\end{itemize}

% ------------------------------- Related Work ---------------------------------------------------
\section{Related Work}
\label{sec:related}
% \subsection{3D Editable Gaussian Splatting}
% The ability to edit 3D Gaussian Splatting (3DGS)~\cite{3DGS} scenes is rapidly emerging. GaussianGrouping~\cite{GaussianGrouping} was a pioneering effort that demonstrated that Gaussians can be grouped into distinct objects for semantic manipulation. Their method relies on lifting 2D segmentation masks generated by models, such as SAM~\cite{SAM}, into three-dimensional (3D) space. Although effective, this 2D-centric projection approach often suffers from view inconsistencies and significant over-segmentation, which is a direct consequence of using 2D masks across multiple views, a limitation that we specifically identified and addressed in this study. Our work differs in that we first construct a 3D-native semantic field, which is then used to actively refine and merge these noisy 2D masks rather than simply projecting them and relying on complex post-processing.

% GaussianEditor~\cite{chen2024gaussianeditor} focuses on the mechanics of manipulation and provides a swift and controllable framework for 3D editing. However, GaussianEditor~\cite{chen2024gaussianeditor} presupposes that a clean, high-quality object selection or mask is available, often relying on user inputs. However, it does not explicitly address the challenge of obtaining robust, multi-view consistent object segmentations from raw two-dimensional (2D) inputs. Thus, our work is complementary, as we focus on providing precise and stable object clusters that are essential for sophisticated editing tools to function effectively.

\subsection{3D Feature Fields and Semantic Distillation} 3D feature fields~\cite{kobayashi2022decomposing} have evolved from NeRF-based distillation~\cite{gsnerf,decomposeFeatureField,n3f} to real-time 3DGS frameworks~\cite{semanticgaussian,zhou2024feature} that embed 2D foundation priors (e.g., CLIP, DINOv2) into Gaussian primitives. 
Recent extensions incorporate hierarchical decomposition~\cite{n2f2}, part-level lifting~\cite{partfield,findanypart}, and 3D-aware prior fine-tuning~\cite{yue2024improving}. 
While some approaches focus on multi-resolution alignment~\cite{LERF,langsplat} or scalability through hash encodings~\cite{legaussian,fmgs}, others improve multiview aggregation for static retrieval~\cite{CF3,semantic_deferred,opengaussian}. 
Unlike these methods, which treat feature fields as static tools, we actively reshape the segmentation by jointly reasoning over semantics, depth, and edges during optimization.

\subsection{Object-Level Scene Understanding and Consistency} 
Bridging the gap between 2D segmentation and 3D object representation remains a fundamental challenge. 
Baselines such as GaussianGrouping~\cite{gaussiangrouping}, SAGA~\cite{segmentany3DGS}, and Splat-SAM~\cite{segthensplat} supervise identity encodings using SAM~\cite{SAM} masks. 
To improve consistency and boundary precision, subsequent works employ joint semantic optimization~\cite{instancegaussian,sai3d,zhang2025bootstraping} or hierarchical clustering~\cite{garfield,superpoint,sagd}. 
Despite these efforts, most methods inherit over-segmentation artifacts of SAM.
In contrast, we treat SAM outputs as an initial over-complete decomposition, refining them through Multi-Cue-Guided Mask Merging (MCM) (Sec.~\ref{sec:multicue_refine}) and cross-view mask matching (Sec.~\ref{sec:cross_view_feature_lift}) to establish globally coherent 3D instances.

\section{Preliminary}
\subsection{3D Gaussian Splatting (3DGS)}
% 3D Gaussian Splatting (3DGS)~\cite{3DGS} is an explicit scene representation that utilizes differentiable 3D Gaussian primitives to model scene.
3D Gaussian Splatting (3DGS)~\cite{3DGS} is an explicit scene representation that utilizes differentiable 3D Gaussian primitives.
Each Gaussian is defined by its mean position $\boldsymbol{\mu}$, opacity $\alpha$, and a 3D covariance matrix $\boldsymbol{\Sigma}$ derived from scaling $\mathbf{s}$ and rotation $\mathbf{q}$. 
The pixel color $C$ is computed through tile-based rasterization and point-based $\alpha$-blending: $C = \sum_{i \in \mathcal{N}} \mathbf{c}_i \alpha_i \prod_{j=1}^{i-1} (1 - \alpha_j)$, where $\alpha_i$ represents the effective opacity at the pixel location. 
This framework can be extended to store high-dimensional feature vectors, forming a 3D feature field for semantic tasks.
\subsection{Segment Anything Model (SAM)}
Segment Anything Model (SAM)~\cite{SAM} is a foundation model for class-agnostic, promptable image segmentation. 
%
% For autonomous scene decomposition, its Automatic Mask Generation (AMG) mode samples dense grid points to produce an exhaustive set of binary masks $\mathcal{M} = \{m_i\}_{i=1}^K$. 
%
For autonomous scene decomposition, its Automatic Mask Generation (AMG) mode produces an initial set of binary masks $\mathcal{M} = \{m_i\}_{i=1}^K$.
While SAM demonstrates exceptional zero-shot generalization, the resulting AMG masks are often 2D-centric and fragmented, lacking cross-view consistency in complex 3D environments.

% ============================================= Method =============================================

\section{Method}
\begin{figure*}[t]
    \centering
    \includegraphics[width=1\linewidth]{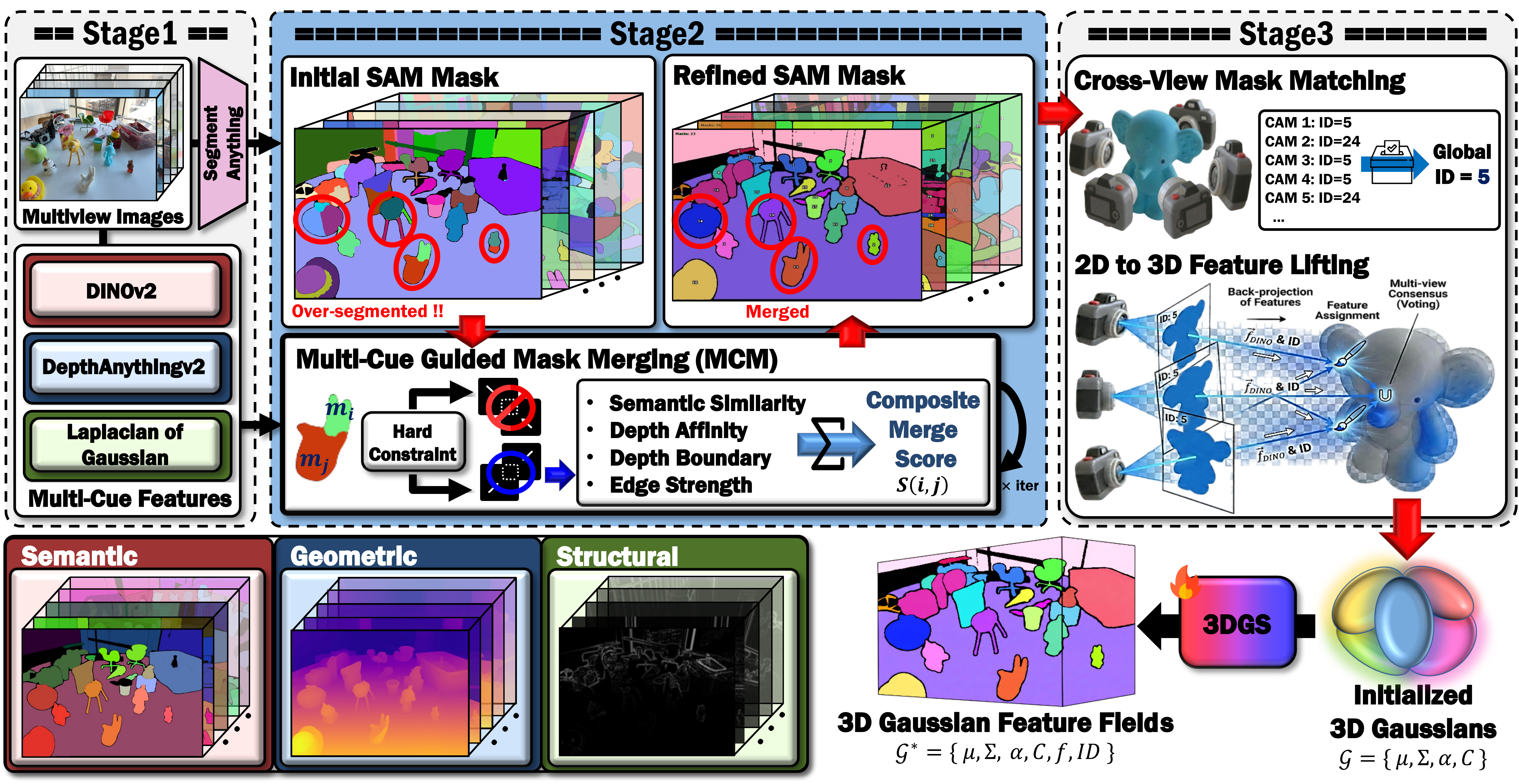}
    \caption{\textbf{Pipeline}. 
    Given multiview images, we extract initial SAM masks and multi-cue priors. 
    These masks are refined via MCM using a composite merge score.
    Following cross-view mask matching, refined 2D features are back-projected and lifted into 3D Gaussians. 
    Consequently, the model outputs a view-consistent object-level feature field for high-fidelity 3D scene understanding.
    }
    \label{fig:pipeline}
\end{figure*}

To address over-segmentation and view inconsistency in 2D-to-3D lifting, we propose a three-stage refinement pipeline (Fig~\ref{fig:pipeline}).
Rather than replacing SAM, we leverage its strength by treating its output as an initial over-complete decomposition that is then refined through semantic, geometric, and structural constraints (Sec.~\ref{sec:multi_cue_features}).
First, we perform Multi-Cue-Guided Mask Merging (MCM) (Sec.~\ref{sec:multicue_refine}) by integrating DINOv2~\cite{dinov2}, monocular depth~\cite{Depthv2}, and LoG edge~\cite{LoG} cues.
This stage consolidates fragmented per-view masks into coherent regions using a scoring framework that balances semantic affinity with geometric boundary preservation.
Second, we establish global object identities through cross-view mask matching and feature lifting (Sec.~\ref{sec:cross_view_feature_lift}).
By utilizing a multiview consensus voting mechanism, we effectively bypass the instability of 2D mask matching and ensure robust identity assignment for each Gaussian.
Finally, a joint optimization (Sec.~\ref{sec:joint_opt}) enforces semantic consistency across the feature field while maintaining the high-fidelity photometric representation inherent to 3DGS.
The comprehensive flow of our pipeline is summarized in Algorithm~\ref{alg:pipeline}.

% -------------------------- Pipeline --------------------------
% --------------------------------------------------------------

% ====================================================================
\subsection{Multi-Cue Feature Extraction}
\label{sec:multi_cue_features}

% 1. Semantic
\noindent\textbf{Semantic Embeddings.} 
We leverage DINOv2~\cite{dinov2} to provide global context that transcends local pixel appearances.
For each candidate mask $m_i$ generated by SAM, where $i$ denotes the mask index, we extract dense feature maps and derive a representative descriptor $f_i$ by averaging the feature vectors across the mask area.
This descriptor serves as a semantic signature for calculating the cosine similarity between candidate masks in~\eqref{eq:composite_score}, enabling the framework to group fragments of the same entity regardless of minor text variations.
%

% 2. Depth
\noindent\textbf{Monocular Depth Estimation.}
We leverage DepthAnythingV2~\cite{Depthv2} to estimate per-pixel relative depth $D \in \mathbb{R}^{H \times W}$, providing geometric context absent from appearance-only segmentation~\cite{SAM}. 
For each mask $m_i$, we compute its mean depth $\bar{d}_i$ and depth gradient magnitude along its boundary $\partial m_i$.
Despite scale ambiguity, this relative depth effectively disambiguates disjoint regions that share similar textures. 
These geometric priors enforce a hard depth constraint and govern the structural penalty terms in~\eqref{eq:composite_score}.

% 3. Edge
\noindent\textbf{Boundary Localization.}
We employ Laplacian-of-Gaussian (LoG)~\cite{LoG} filtering to produce an edge-strength map $\nabla_{\text{edge}}$ for structural boundary localization.
This rotation-invariant prior localizes structural discontinuities, ensuring precise alignment with object boundaries across diverse orientations.
As a penalty term in~\eqref{eq:composite_score}, $\nabla_{\text{edge}}$ discourages the erroneous merging of distinct objects along their shared boundary $\partial m_{ij}$, even when they exhibit high semantic similarity.
\subsection{Multi-Cue-Guided Mask Merging (MCM)}
\label{sec:multicue_refine}
% To resolve over-segmentation of SAM, we address the inherent limitations of semantic-only merging. 
%
To resolve over-segmentation, we address the inherent limitations of semantic-only merging.
These failures typically involve insufficient merging of noisy textures or erroneous grouping of visually similar objects located at distinct depths. 
To reduce these issues, we first utilize a hard depth constraint to exclude candidate pairs located at significantly different depths.
Merge decisions for the remaining candidates are then decided by a composite merge score that synthesizes semantic and structural cues.

\noindent\textbf{Adjacency Computation.}
To focus the merging process on plausible candidates, we define two masks $m_i$ and $m_j$, where $i$ and $j$ denote the respective mask indices, as adjacent if they satisfy three geometric criteria: 
(1) Bounding box proximity within $\tau_{\text{gap}}$ pixels. 
(2) Intersection following morphological dilation such that $\text{dilate}(m_i) \cap \text{dilate}(m_j) \neq \emptyset$; 
and (3) Contact ratio exceeding $\tau_{\text{contact}}$. 
This pruning strategy effectively eliminates over 80\% of irrelevant candidate pairs, significantly reducing computational overhead while preserving all geometrically plausible merges.\\
\noindent\textbf{Hard Depth Constraint.}
%For these adjacent candidates, we first apply a hard geometric constraint to prevent erroneous merges across foreground–background boundaries by enforcing $|\bar{d}_i - \bar{d}_j| < \tau_{\text{depth}}\label{eq:hard_depth}$ and depths are normalized to $[0,1]$.
%
For these adjacent candidates, we first apply a hard geometric constraint to prevent erroneous merges across depth discontinuities by enforcing $|\bar{d}_i - \bar{d}_j| < \tau_{\text{depth}}\label{eq:hard_depth}$ and depths are normalized to $[0,1]$.
This hard constraint ensures foreground-background separation even when semantic features align (e.g., identical textures at different depths), resolving the white-on-white dilemma where appearance-only methods fail.

% -------------------------
\noindent\textbf{Composite Merge Score.}
For adjacent pairs that satisfy the hard depth constraint, we compute a unified composite score to determine the final merge.
This score synthesizes semantic similarity with structural and geometric penalties to ensure boundary-aware consolidation as follows:
\begin{equation}
\label{eq:composite_score}
\begin{aligned}
    S(i,j) = \,\, &w_{\text{sem}} \cdot \cos(f_i, f_j) + w_{\text{depth}} \cdot \exp\!\left( - \frac{|\bar{d}_i - \bar{d}_j|^2}{2\sigma_d^2} \right) \\
    & - w_{\text{dbound}} \cdot \max(\nabla_D \cap \partial m_{ij}) - w_{\text{edge}} \cdot \max(\nabla_{\text{edge}} \cap \partial m_{ij}),
\end{aligned}
\end{equation}
where $S(i,j)$ evaluates the affinity between masks $i$ and $j$.
In this formulation, $\nabla_{\text{edge}}$ denotes the LoG edge magnitude, $\nabla_D$ represents the depth gradient, and $\partial m_{ij} = \partial m_i \cup \partial m_j$ represents the union of region boundaries.
The composite merge score integrates four terms to balance semantic unification with structural preservation. 
First, the semantic term leverages high-dimensional DINOv2 embeddings to group fragments that belong to the same entity.
Second, the depth affinity term promotes the unification of coplanar segments by evaluating their geometric continuity in 3D space.
Third, the depth boundary term penalizes merges across sharp depth transitions to maintain separation between foreground and background objects.
Finally, the edge magnitude term enforces structural integrity by preventing the consolidation of distinct objects separated by strong visual boundaries.
%
% Through this synthesis, $S(i,j)$ optimizes the trade-off between recall and precision during the consolidation process.
Through this synthesis, $S(i,j)$ optimizes the trade-off between recall and precision during the process.
The framework executes a merge only when a mask pair satisfies the hard depth constraint and exceeds the threshold $\tau_{\text{merge}}$.
Under this protocol, pairs failing the initial constraint are immediately classified as distinct entities and excluded from further evaluation.
This ensures that only spatially plausible candidates proceed to the final assessment, where their structural and semantic coherence is verified for unification.

\noindent\textbf{Iterative Refinement.}
We apply greedy merging iteratively (max 1,000 iterations), recomputing adjacency and scores after each merge. The process terminates when no pair satisfies all constraints.
%

% Algorithm
\begin{algorithm}[t]
\caption{Overview of Proposed Refinement Pipeline}
\label{alg:pipeline}
\begin{algorithmic}[1]
    \REQUIRE Images $\{I_n\}_{n=1}^N$, Initial Masks $\{M_n\}_{n=1}^N$, 3D Gaussians $\mathcal{G}$
    \ENSURE Feature-enriched 3D Gaussians $\mathcal{G}_{feat}$
    
    \STATE \textbf{Stage I: Multi-Cue Feature Construction} ~~~~~~~~~~~~~~~~~~~~~~~~~~// Sec.~\ref{sec:multi_cue_features}
    \FOR{each view $n \in \{1, \dots, N\}$}
        \STATE Extract DINOv2 ($f$), LoG Edges ($\nabla_{\text{edge}}$), and Depth ($D$)
    \ENDFOR
    
    \medskip
    \STATE \textbf{Stage II: Multi-Cue-Guided Mask Merging} ~~~~~~~~~~~~~~~~~~~ // Sec.~\ref{sec:multicue_refine}
    \FOR{each view $n$}
        \REPEAT
            \STATE Identify adjacent masks $m_i, m_j$
            \IF{$|\bar{d}_i - \bar{d}_j| < \tau_{depth}$} 
                \STATE Compute composite merge score $S(i,j)$ via \eqref{eq:composite_score} 
                \IF{$S(i,j) > \tau_{merge}$}
                    \STATE Update $m_{new} \leftarrow \text{Merge}(m_i, m_j)$
                \ENDIF
            \ENDIF
        \UNTIL{convergence}
    \ENDFOR
    
    \medskip
    \STATE \textbf{Stage III: Cross-view Mask Matching and 3D Lifting} ~~~~~~ // Sec.~\ref{sec:cross_view_feature_lift}
    \STATE $\mathcal{C}_{global} \leftarrow \text{GlobalClustering}(\{m_n\})$ via DINO similarity
    \FOR{each Gaussian $g \in \mathcal{G}$}
        \STATE Project $g$ onto visible camera views $n \in \mathcal{V}_g$
        \STATE $ID_g \leftarrow \text{MajorityVote}(\text{Global IDs from projected views})$ 
        \STATE $\sigma^2_{feat} \leftarrow \text{ComputeVariance}(g, \mathcal{V}_g)$ 
        \IF{$\sigma^2_{feat} > \tau_{var}$} 
            \STATE Filter $g$ (Reliability check)
        \ENDIF
    \ENDFOR
    
    \medskip
    \STATE \textbf{return} Final optimized feature field $\mathcal{G}_{feat}$
\end{algorithmic}%
\end{algorithm}

% ====================================================================
\subsection{Cross-View Mask Matching \& Feature Lifting to 3D Gaussians}
\label{sec:cross_view_feature_lift}

\noindent\textbf{Cross-View Mask Matching.}
The preceding stage produces geometrically coherent but viewpoint-specific mask regions.
To establish global consistency, we perform Global Mask Association by constructing an affinity matrix from the representative descriptors $\bar{f}_i$, representing the mean features of consolidated masks.
Two masks $m_i$ and $m_j$ from different viewpoints are assigned a shared global identity if their cosine similarity exceeds the matching threshold $\tau_{\text{match}}$:

\begin{equation} 
    \cos(\bar{f}_i, \bar{f}_j) > \tau_{\text{match}}. 
\end{equation} 

\noindent This deterministic association enables the GlobalClustering of masks across the entire scene, ensuring that unique object instances maintain stable identities throughout the reconstructed scene.

\noindent\textbf{Feature Lifting to 3D Gaussians.}
Once global identities are established across all viewpoints, we lift these 2D semantic assignments into a 3D Gaussian representation to form a consistent feature field.
This process bridges the gap between viewpoint-specific 2D masks and the underlying 3D structure by populating each Gaussian $g$ with a representative identity feature, $f_g$.
Distinct from the 2D semantic descriptor $f_i$, the identity feature $f_g$ serves as a high-dimensional 3D signature that anchors global object consistency across the Gaussian field.

\noindent\textbf{Multi-View Semantic Fusion.}
For each 3D Gaussian $g$ with center position $\mu_g$, we aggregate information from all $N$ cameras through a weighted fusion of the corresponding 2D features:

\begin{equation}
    f_g = \frac{\sum_{v \in \mathcal{V}g} w_v \cdot f_{m_v(g)}}{\sum_{v\in \mathcal{V}g} w_v}
\end{equation}

\noindent where $\mathcal{V}g$ is the set of viewpoints where Gaussian $g$ is visible, aggregation weight $w_v$, and $f_{m_v(g)}$ is the refined identity feature of the mask $m_v(g)$.
This weighted scheme prioritizes viewpoints with clear, orthogonal visibility, thereby minimizing the impact of perspective distortion and partial occlusions.
The resulting feature $f_g$ serves as a robust 3D descriptor. enabling the subsequent assignment of stable object identities through the multiview consensus mechanism.

\noindent\textbf{3D Mask ID Assignment.}
To elucidate object representation, each Gaussian is assigned a definitive label using a multiview majority voting mechanism.
The assigned identity $\text{ID}_g$ is determined by identifying the most frequent global ID across all the visible viewpoints:

\begin{equation}
    \text{ID}_g = \arg\max_{k} \sum_{v \in \mathcal{V}_g} \bbbone[\text{ID}_{m_v(g)} = k],
\end{equation}

\noindent where $\text{ID}_{m_v(g)}$ is the global identity of the mask containing the projection of Gaussian $g$ in view $v$, and $\bbbone[\cdot]$ denotes the indicator function.
This voting strategy effectively filters out transient single-view noise and misalignments, ensuring that each 3D primitive is assigned to the most statistically reliable region.
By establishing a deterministic mapping between global IDs and their respective color encodings, this mechanism ensures that refined masks exhibit identical identity values and consistent color signatures across all viewpoints.

\noindent\textbf{Variance-based Reliability Filtering.}
To ensure that only consistent multiview evidence is used, we perform a statistical validation of the lifted features.
For each 3D Gaussian $g$, we compute the feature variance $\sigma^2_{\text{feat}}(f_g)$ across all views where the primitive is visible:

\begin{equation}
    \sigma^2_{feat}(f_g) = \frac{1}{|\mathcal{V}_g|} \sum_{v \in \mathcal{V}_g} \| f_{m_v(g)} - f_g \|^2.
    \label{eq:var-filtering}
\end{equation}

\noindent where $\mathcal{V}_g$ is the set of viewpoints in which gaussian $g$ satisfies the visibility criteria, $|\mathcal{V}_g|$ is the total number of visible views, and $f_{m_v(g)}$ represents the 2D identity feature projected from view $v$.
Gaussians with $\sigma^2_{feat}(f_g) > \tau_{\text{var}}$ are discarded as unreliable.
Such high-variance assignments typically arise from occlusion boundaries or inconsistent 2D segmentation across different viewpoints.
This filtering ensures that the noise inherent in the single-view model outputs is reduced.

% ====================================================================
\subsection{Joint Optimization}
\label{sec:joint_opt}
The 3D representation is optimized end-to-end using a composite loss function:
\begin{equation}
\mathcal{L} = \mathcal{L}_\text{render} + \lambda_{s}\mathcal{L}_\text{semantic} + \lambda_\text{reg}\mathcal{L}_\text{3D-reg},
\end{equation}
where $\mathcal{L}_\text{render}$ represents the photometric rendering loss following standard 3DGS~\cite{3DGS}, $\mathcal{L}_\text{semantic}$ denotes the identity consistency loss, and $\mathcal{L}_\text{3D-reg}$ signifies the structural 3D regularization term, with $\lambda_s$ and $\lambda_\text{reg}$ serving as balancing weights. 
%
% Following 3DGS~\cite{3DGS}, $\mathcal{L}_\text{render}$ consists of a weighted sum of $\mathcal{L}_1$ and $\mathcal{L}_\text{SSIM}$ to ensure photometric fidelity.
%

\noindent\textbf{Semantic Cohesion Loss.}
To ensure that the 3D representation inherits the refined 2D semantics, we attach learnable semantic features to each Gaussian and supervise them using the lifted features.

\begin{equation}
    \mathcal{L}_{\text{semantic}} = \frac{1}{|\mathcal{G}_{\text{valid}}|} \sum_{g \in \mathcal{G}_{\text{valid}}} \|f_g - \tilde{f}_g\|^2,
    \label{loss:semantic}
\end{equation}

\noindent where $\tilde{f}_g$ is the target lifted feature and $\mathcal{G}_{\text{valid}}$ denotes the subset of Gaussians with a reliable multiview consensus.
Although semantic features are optimized jointly with geometry and appearance, this loss is applied only after the completion of the densification process to prevent semantic gradients from interfering with the initial geometric convergence.

\noindent\textbf{3D Geometric Regularization.}
We encourage spatial smoothness via k-NN consistency~\cite{knn} as follows:

\begin{equation}
\mathcal{L}_{\text{3D-reg}} = \frac{1}{|\mathcal{G}|} \sum_{g \in \mathcal{G}} \sum_{g' \in \mathcal{N}_5(g)} \|f_g - f_{g'}\|^2,
\end{equation}

\noindent where $\mathcal{N}_5(g)$ denotes 5 nearest spatial neighbors. 
This regularization produces coherent semantic fields that are suitable for object-level manipulation.

\noindent\textbf{Two-Stage Optimization.}
We adopt two-stage training schedule: 
(1) Stage 1 (iter 0--15k): Standard 3DGS optimization focusing on geometry and appearance, with semantic features frozen. 
(2) Stage 2 (iter 15k--100k): Joint optimization of geometry, appearance, and semantic features with $\mathcal{L}_{\text{semantic}}$ and $\mathcal{L}_{\text{3D-reg}}$ activated. 
This schedule stabilizes early reconstruction while ensuring convergence to semantically coherent representations

% ========================================= Experiments ============================================
% ==================================================================================================
\section{Experiments}
\subsection{Experimental Settings}

We implement our proposed framework based on the official codebase of GaussianGrouping~\cite{gaussiangrouping}. 
All 3DGS~\cite{3DGS} parameters and feature fields are optimized for 30,000 iterations on a single RTX 3090 GPU.
%
% 
% Extracting multi-cue maps takes approximately 10 hours, while subsequent joint optimization requires about 2 hours. 
% %
% Although this adds one-time offline overhead, it significantly enhances the reliability of the resulting 3D primitives.
% }
%
For evaluation, we utilize the LERF~\cite{LERF}, Replica~\cite{straub2019replica}, and in-the-wild datasets from~\cite{lee2024consistent} to validate robustness across diverse real-world environments.
The rendering quality is assessed via PSNR, SSIM~\cite{psnr_ssim}, and LPIPS~\cite{lpips} to ensure that our joint optimization preserves a high-fidelity photometric representation while regularizing the feature field. 
Segmentation performance is measured through mIoU and Mask mIoU (mMIoU), while Boundary mIoU (mBIoU) specifically evaluates mask alignment with physical boundaries for actionable primitives. 
Finally, we report the average mask count (\#Mask) per scene to quantify the capacity of our framework to reduce over-segmentation in the raw foundation model outputs.

\noindent\textbf{Hyperparameters.} 
For mask merging in~\eqref{eq:composite_score}, we empirically set weights to $w_{sem}=0.5$, $w_{edge}=0.2$, $w_{depth}=0.2$, and $w_{dbound}=0.1$ with $\sigma_d^2=0.05$.
To satisfy the dual condition for merging, we define two key thresholds: the merging threshold $\tau_{merge}=0.85$, and the hard constraint $\tau_{depth}=0.15$. 
While $\tau_{\text{depth}}$ acts as a hard constraint to filter out disjoint regions, $\tau_\text{merge}$ determines the final merge based on the composite merge score.
For cross-view matching, we set the similarity threshold $\tau_{match}=0.7$, and a variance threshold $\tau_{var}=0.05$ in~\eqref{eq:var-filtering}. 
These values remain fixed across all datasets to demonstrate the robustness of our framework.

% ---------------------------------------------

\subsection{Qualitative Results}
\begin{figure}[t]
    \centering
    \includegraphics[width=1\linewidth]{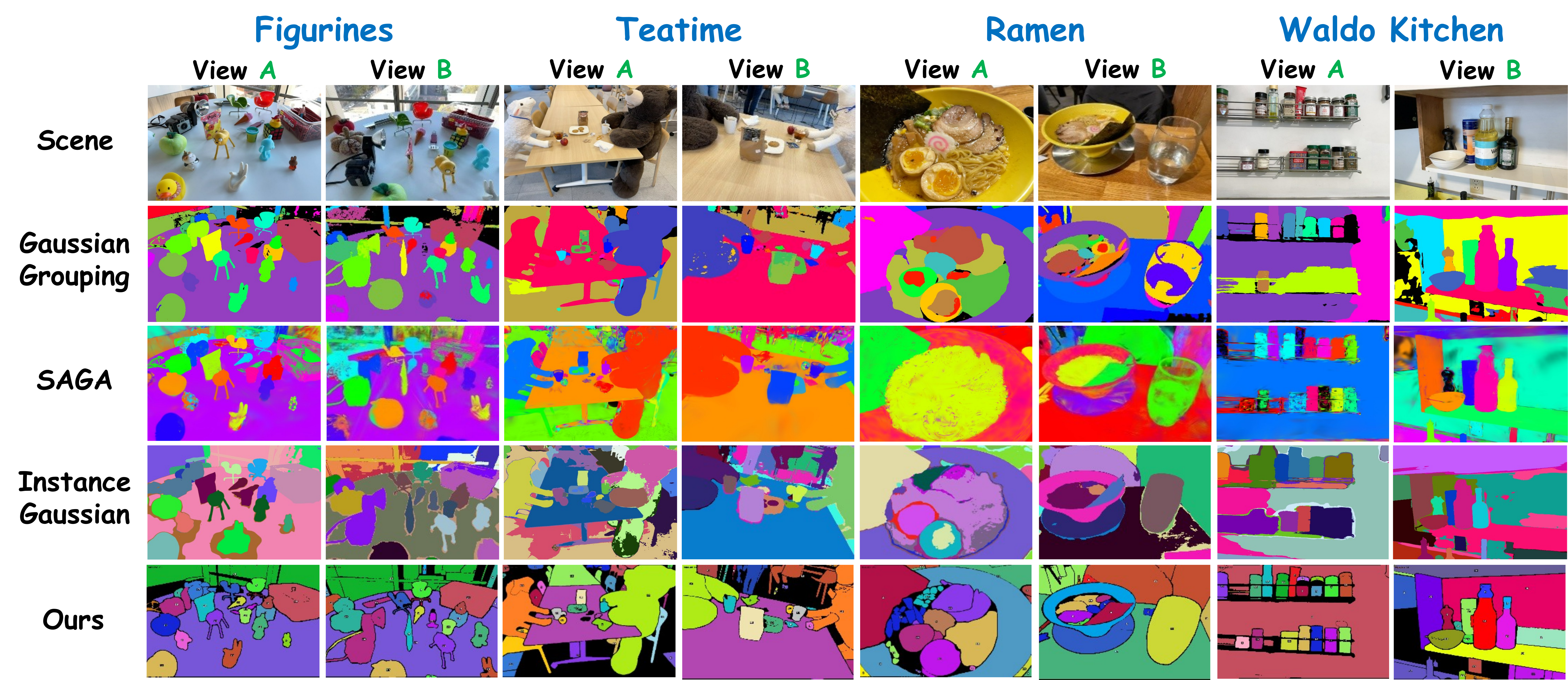}
    \caption{\textbf{Qualitative Result.}
    This figure demonstrates how faithfully our results represent the scene and maintain view consistency across different viewpoints.}
    \label{fig:qualitative results}
\end{figure}

We qualitatively compare our method against state-of-the-art (SOTA) 3D segmentation frameworks~\cite{gaussiangrouping,segmentany3DGS,instancegaussian}.
As illustrated in Fig.~\ref{fig:qualitative results}, while baseline methods achieve reasonable object localization, they consistently struggle with identity fragmentation and boundary leakage. 
%
% Specifically, GaussianGrouping tends to produce over-segmented results, such as a ramen bowl or teddy bear. 
Specifically, GaussianGrouping tends to produce over-segmented results.
This fragmentation occurs because these methods directly inherit the raw output of SAM without accounting for the underlying 3D geometric continuity.
Furthermore, InstanceGaussian occasionally suffers from "\textit{label bleeding}" where object boundaries become blurred or misaligned in complex scenes with overlapping depth layers.
In contrast, our approach yields significantly cleaner and more stable object masks by employing MCM (Sec.~\ref{sec:multicue_refine}). 
By evaluating fragmented regions through scoring mechanism, we successfully unify texture-induced fragments.

% ---------------------------------------------------------------------------------------------------------------------

%------------------ Experiment1 ------------------
\begin{table}[t]
    \caption{\textbf{Quantitative Comparison on LERF and Replica Dataset.} \\The \textbf{bold} score is the best, and the \underline{underline} score is the second.}
    \centering
    \resizebox{\columnwidth}{!}{%
    \begin{tabular}{llcccccc}
        \toprule
        Category & Model & PSNR $\uparrow$ & SSIM $\uparrow$ & LPIPS $\downarrow$ & mIoU $\uparrow$ & mBIoU $\uparrow$ & \#Mask $\downarrow$ \\ 
        \midrule
        Standard & 3DGS~\cite{3DGS} & \underline{28.6} & \underline{0.88} & \textbf{0.11} & - & - & - \\
        \midrule
        \multirow{3}{*}{\shortstack{3D\\Segmentation}} 
        & Gau-Group~\cite{gaussiangrouping} & \underline{28.6} & \underline{0.88} & \underline{0.12} & 0.685 & 0.642 & 9,627 \\
        & SAGA~\cite{segmentany3DGS} & 28.2 & 0.87 & 0.15 & \underline{0.712} & 0.658 & 178 \\
        & InstanceGaussian~\cite{instancegaussian} & \textbf{28.8} & 0.87 & 0.15 & 0.702 & \underline{0.662} & 101 \\
        \midrule
        \multirow{3}{*}{\shortstack{Feature Field}} 
        & Feature3DGS~\cite{zhou2024feature} & 27.5 & 0.84 & 0.18 & 0.538 & 0.485 & - \\
        & GARField~\cite{garfield} & 28.1 & 0.87 & 0.16 & 0.710 & 0.615 & - \\
        & CF3~\cite{CF3} & 28.3 & 0.87 & 0.14 & 0.524 & 0.582 & - \\
        \midrule
        \textbf{Proposed} & \textbf{Ours} & \underline{28.6} & \textbf{0.89} & 0.13 & \textbf{0.728} & \textbf{0.677} & \textbf{67} \\ 
        \bottomrule
    \end{tabular}%
    }
    \label{tab:exp1}
\end{table}
% ------------------------------------------------

\subsection{Quantitative Results}

Table~\ref{tab:exp1} summarizes the quantitative performance of our method compared to vanilla 3DGS~\cite{3DGS}, SOTA 3D segmentation~\cite{segmentany3DGS,instancegaussian,gaussiangrouping} and feature field based approaches~\cite{garfield,CF3,zhou2024feature} on the LERF and Replica datasets.\\
\noindent\textbf{Photometric Rendering Quality.} As shown in Table~\ref{tab:exp1}, our method maintains high-fidelity rendering performance comparable to the vanilla 3DGS. 
While InstanceGaussian achieves the highest PSNR of 28.8, our framework matches the baseline at 28.6 and achieves a competitive SSIM of 0.89. 
This indicates that our joint optimization and mask merging process does not compromise the underlying photometric reconstruction quality of the 3D Gaussian representations.\\
\noindent\textbf{Segmentation Accuracy.} In terms of 3D segmentation, our method achieves SOTA performance across all primary metrics. 
We report a Mean IoU (mIoU) of 0.728, outperforming the second-best method, SAGA~\cite{segmentany3DGS} (0.712), and significantly surpassing feature field-based approaches like Feature3DGS~\cite{zhou2024feature} (0.538). 
Furthermore, our method excels in boundary precision, reaching a Mean Boundary IoU (mBIoU) of 0.677. 
This improvement highlights the effectiveness of our multi-cue integration, particularly the use of LoG-based structural priors and depth-aware constraints, in producing sharp and physically accurate object boundaries compared with existing distilled feature field methods.\\
\noindent\textbf{Resolution of Over-segmentation.} The most significant contribution of our framework is the drastic reduction in the over-segmentation inherent in foundation model priors. 
While GaussianGrouping generates an average of 9,627 masks per scene due to the fragmented nature of SAM outputs, our strategy successfully merges them into 67 masks. 
Even compared to other models like SAGA and InstanceGaussian, our method produces the most compact and semantically consistent object representations. 
This reduction confirms that our approach effectively extracts "\textit{actionable}" object primitives suitable for downstream tasks.

\subsection{Analysis}

\begin{figure*}[t]
    \centering
    \includegraphics[width=1\linewidth]{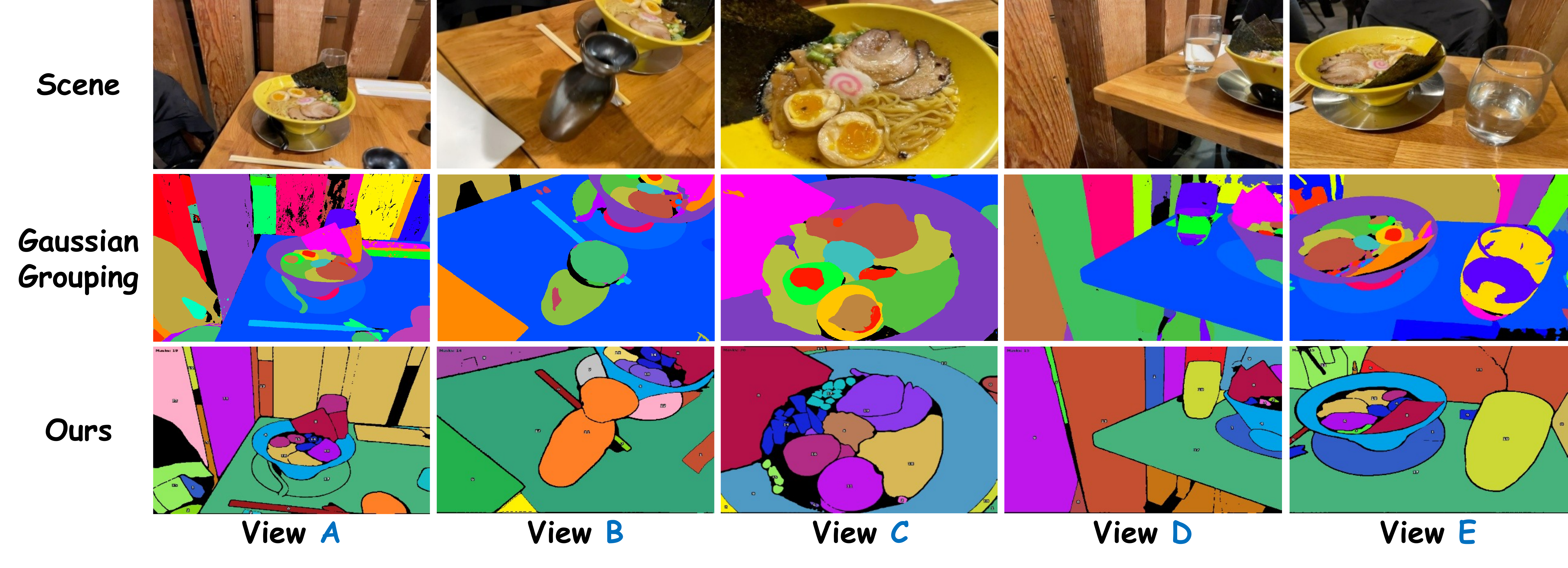}
    \caption{\textbf{Qualitative Comparison of View Consistency.}
    Compared to the baseline~\cite{gaussiangrouping}, which exhibits over-segmentation and inconsistent object identities across different views, our approach successfully produces coherent, view-consistent object masks across changing viewpoints.}
    \label{fig:consistency_comparison}
\end{figure*}

\noindent\textbf{View Consistency.}
We qualitatively evaluate cross-view consistency using the "Ramen" scene. 
%
% While the baseline~\cite{gaussiangrouping} inherits over-segmentation of SAM and produces fluctuating identities across different views, our framework consolidates these fragments through Sec.~\ref{sec:multicue_refine}. 
%
While the baseline~\cite{gaussiangrouping} inherits over-segmentation of SAM and produces fluctuating identities across different views, our framework merges these fragments through Sec.~\ref{sec:multicue_refine}. 
As illustrated in Fig.~\ref{fig:consistency_comparison} (View A--E), our model maintains stable object identities and sharp boundaries despite significant viewpoint changes.
These results demonstrate that our refined masks serve as robust actionable primitives, successfully bridging the gap between fragmented 2D priors and coherent 3D object representations.

% -----------------------------------------------------------------------------------------------
\begin{table*}[t]
    \caption{\textbf{Ablation Studies on LERF Figurine Dataset.} This table quantifies the incremental performance gains achieved by sequentially integrating semantic~\cite{dinov2}, geometric~\cite{Depthv2}, and structural~\cite{LoG} priors into our framework.}
    \label{tab:Multicue}
    \centering
    \small
    \begin{tabular}{l ccc ccccc}
        \toprule
        \multirow{2}{*}{Configuration} & \multicolumn{3}{c}{Signals} & \multicolumn{5}{c}{Metrics} \\
        \cmidrule(r){2-4} \cmidrule(l){5-9}
        & DINOv2 & LoG & DAV2 & mIoU $\uparrow$ & \# Masks $\downarrow$ & PSNR $\uparrow$ & SSIM $\uparrow$ & LPIPS $\downarrow$ \\
        \midrule
        Baseline~\cite{gaussiangrouping} & -- & -- & -- & 0.685 & 9,627 & \textbf{28.6} & 0.88 & 0.13 \\
        Ours (1) & \checkmark & -- & -- & 0.721 & 245 & 28.4 & 0.89 & 0.14 \\
        Ours (2) & \checkmark & \checkmark & -- & 0.729 & 112 & 28.5 & \textbf{0.91} & \textbf{0.12} \\
        \midrule
        \textbf{Ours (Full)} & \checkmark & \checkmark & \checkmark & \textbf{0.741} & \textbf{67} & \textbf{28.6} & \textbf{0.91} & \textbf{0.12} \\
        \bottomrule
        \multicolumn{9}{l}{\footnotesize * DAV2 denotes DepthAnythingV2.}
    \end{tabular}
\end{table*}

\noindent\textbf{Impact of Multi-Cue Priors.} 
Table~\ref{tab:Multicue} quantifies the incremental gains from integrating semantic, structural, and geometric priors. 
The baseline~\cite{gaussiangrouping} exhibits extreme fragmentation with 9,627 masks, illustrating the pitfalls of lifting raw SAM outputs directly into 3D spaces. 
Adding DINOv2 semantic priors (Ours (1)) reduces the mask count to 245, proving that high-level context is essential for initial object consolidation. 
Incorporating structural priors via LoG (Ours (2)) further refines these groupings with sharp edge guidance and improves rendering metrics through precise boundary delineation. 
Our full model, which includes DepthAnythingV2 geometric priors, achieves state-of-the-art performance with 0.741 mIoU and a consolidated set of 67 actionable primitives.
%

% --------------------

\begin{figure}[t]
    \centering
    \includegraphics[width=0.95\linewidth]{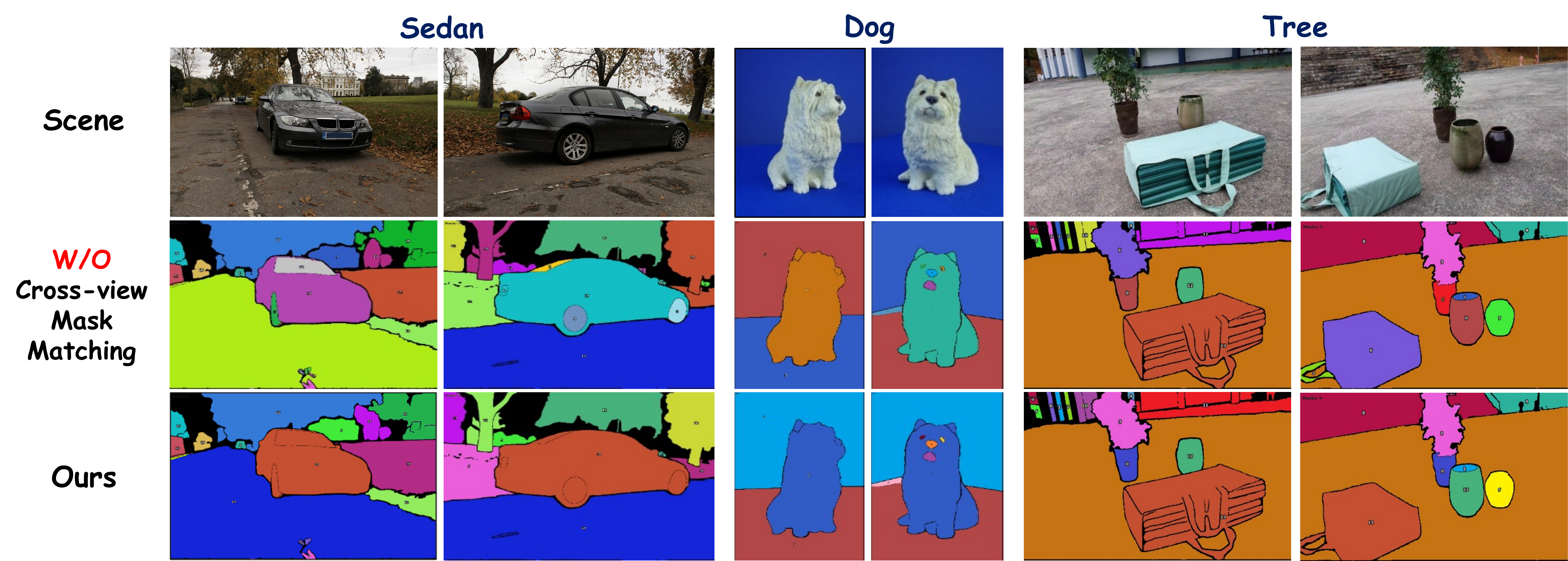}
    \caption{\textbf{Qualitative Ablation of Cross-View Mask Matching.}
    Without cross-view matching (middle row), masks are restricted to per-view assignments, leading to inconsistent identities.
    Our full pipeline (bottom row) ensures global identity coherence, which is essential for stable 3D representation.}
    \label{fig:ablation1}
\end{figure}

\noindent\textbf{Impact of Cross-View Mask Matching.}
To validate cross-view mask matching (Sec.~\ref{sec:cross_view_feature_lift}), we compare our framework against a baseline that refines masks only within individual views.
As shown in Fig.~\ref{fig:ablation1}, per-view refinement resolves local over-segmentation but fails to maintain identity consistency across different viewpoints. 
This lack of coherence results in identity flickering, where a single object is assigned different IDs across viewpoints. 
Such instability prevents our mechanism (Sec.~\ref{sec:cross_view_feature_lift}) from effectively aggregating features into coherent 3D primitives. 
Our strategy resolves this issue by establishing stable correspondences through global semantic affinities. 
This approach ensures consistent object assignment and transforms the segments into coherent actionable primitives.

% ============================================= Conclusion =========================================
\section{Conclusion}
We present a unified framework for 3D scene decomposition that transforms fragmented 2D priors into structured and view-consistent object primitives. 
By treating the foundation model outputs as an initial overcomplete state, our approach effectively bridges the gap between per-view 2D perception and coherent 3D representation. 
The proposed Multi-Cue-Guided Mask Merging process addresses over-segmentation of SAM through multi-cue integration, while cross-view mask matching and feature lifting ensure global identity coherence. 
These components enable the extraction of high-fidelity primitives that maintain geometric and semantic consistency across viewpoints, providing a robust foundation for structured 3D scene understanding and downstream applications.

% --------------------------------------------- References ------------------------------------------

% --- cleveref 약어(abbreviation) 수동 설정 ---
% \Cref (문장 시작)용
\Crefname{equation}{Eq.}{Eqs.}
\Crefname{figure}{Fig.}{Figs.}
\Crefname{table}{Tab.}{Tabs.}
\Crefname{section}{Sec.}{Secs.}

% \cref (문장 중간)용
\crefname{equation}{eq.}{eqs.}
\crefname{figure}{fig.}{figs.}
\crefname{table}{tab.}{tabs.}
\crefname{section}{sec.}{secs.}
% ------------------------------------------------

\section{Implementation Details}
\label{sec:impl_detail}

\subsection{Network Architecture}
\noindent\textbf{DINO Feature Extraction.}
We employ DINOv2~\cite{dinov2} (ViT-B/14) as the pretrained backbone for semantic feature extraction, leveraging its discriminative representation for robust object-level understanding.
Input images are resized to $518 \times 518$ pixels, ensuring precise alignment with the $14 \times 14$ patch size.
The model yields 768-dimensional per-patch features from the final layer, capturing dense semantic signatures even in cluttered environments.
To facilitate stable cosine similarity calculations, all extracted descriptors are $L_2$-normalized before the Multi-Cue-Guided Mask Merging (MCM) process.

\noindent\textbf{Object Feature Field.} 
Each 3D Gaussian is augmented with a learnable 16-dimensional object feature vector $\mathbf{o} \in \mathbb{R}^{16}$, which is initialized from a zero-centered Gaussian distribution $\mathcal{N}(0, 0.01)$. 
These features are jointly optimized with the Gaussian geometry and appearance parameters to enable consistent object-level grouping. 
During rendering, the object features are splatted via the standard 3DGS rasterization pipeline, producing 16-channel identity feature maps for 3D-to-2D supervision.

\subsection{Optimization Details}
\noindent\textbf{Learning Rates.} 
We employ the Adam optimizer for all parameters. 
The learning rates for each parameter type are as follows:
\begin{itemize}
    \item \textbf{Position (xyz)}: Exponential decay from $1.6 \times 10^{-4}$ to $1.6 \times 10^{-6}$ over 50k iterations.
    \item \textbf{Object features}: $2.5 \times 10^{-3}$ (fixed during Stage 1).
    \item \textbf{Opacity}: $5.0 \times 10^{-2}$.
    \item \textbf{Scaling}: $5.0 \times 10^{-3}$.
    \item \textbf{Rotation}: $1.0 \times 10^{-3}$.
\end{itemize}

\noindent\textbf{Training Schedule.} 
The total optimization process takes 100k iterations on a single NVIDIA RTX 3090 GPU. 
Densification and pruning are performed every 100 iterations between 500 and 15k iterations. 
To ensure that the semantic features do not destabilize the geometric convergence, we freeze the 16-dimensional object features during the first 15k iterations. 
After the densification phase, we activate the joint optimization with $\lambda_s = 0.1$ and $\lambda_{reg} = 0.01$.
%

% ------------------------------------------------------------------------------------------------------
\section{Model Selection}
\label{sec:Model Selection}

\begin{table}[t]
    \centering
    \caption{\textbf{Quantitative Analysis of Semantic Backbones.} We assess the discriminative power of various priors using Intra-object and Inter-object cosine similarities. The Separation Ratio ($\uparrow$) represents the relative gap between internal consistency and external distinctness.}
    \label{tab:feature_analysis}
    \small
    \begin{tabular}{lcccc}
        \toprule
        \textbf{Feature} & \textbf{Intra $\uparrow$} & \textbf{Inter $\downarrow$} & \textbf{Ratio $\uparrow$} & \textbf{Dim} \\
        \midrule
        CLIP-ViT-B/16  & 0.71 & 0.43 & 1.65 & 512 \\
        SAM mask feat. & 0.68 & 0.51 & 1.33 & 256 \\
        DINOv2-Small~\cite{dinov2}   & 0.84 & 0.29 & 2.90 & 384 \\
        \textbf{DINOv2-Base~\cite{dinov2}} & \textbf{0.89} & \textbf{0.21} & \textbf{4.24} & 768 \\
        \bottomrule
    \end{tabular}
\end{table}

Our framework integrates three distinct foundation models to achieve robust multi-cue feature extraction. 
Here, we provide an extended justification for these architectural decisions, supported by theoretical analysis and empirical observations.

\subsection{Monocular Depth Estimation: DepthAnythingv2}
We employ DepthAnythingv2~\cite{Depthv2} as our primary geometric prior. 
Unlike earlier monocular depth estimators such as MiDaS~\cite{monoculardepth} or DPT~\cite{DPT}, which often produce over-smoothed depth maps, Depth Anything V2 exhibits an exceptional boundary sharpness. 
This is critical for our framework because our hard depth constraint relies on detecting sharp discontinuities to prevent "bleeding" between foreground objects and the background.
Furthermore, training the model on a massive 62M image dataset provides a level of zero-shot robustness that is essential for diverse 3DGS captures (e.g., Mip-NeRF 360~\cite{MipNeRF360}, LERF~\cite{LERF}). 
In our experiments, we observed that while appearance-only features (DINOv2) often suffer from visual aliasing, where two white walls at different depths produce identical semantic embeddings, Depth Anything V2 successfully resolves this ambiguity. 
By providing the missing geometric context, it allows the system to treat these walls as physically disjoint entities, effectively solving the "white-on-white dilemma."

\begin{figure}[t]
    \centering
    \includegraphics[width=1\linewidth]{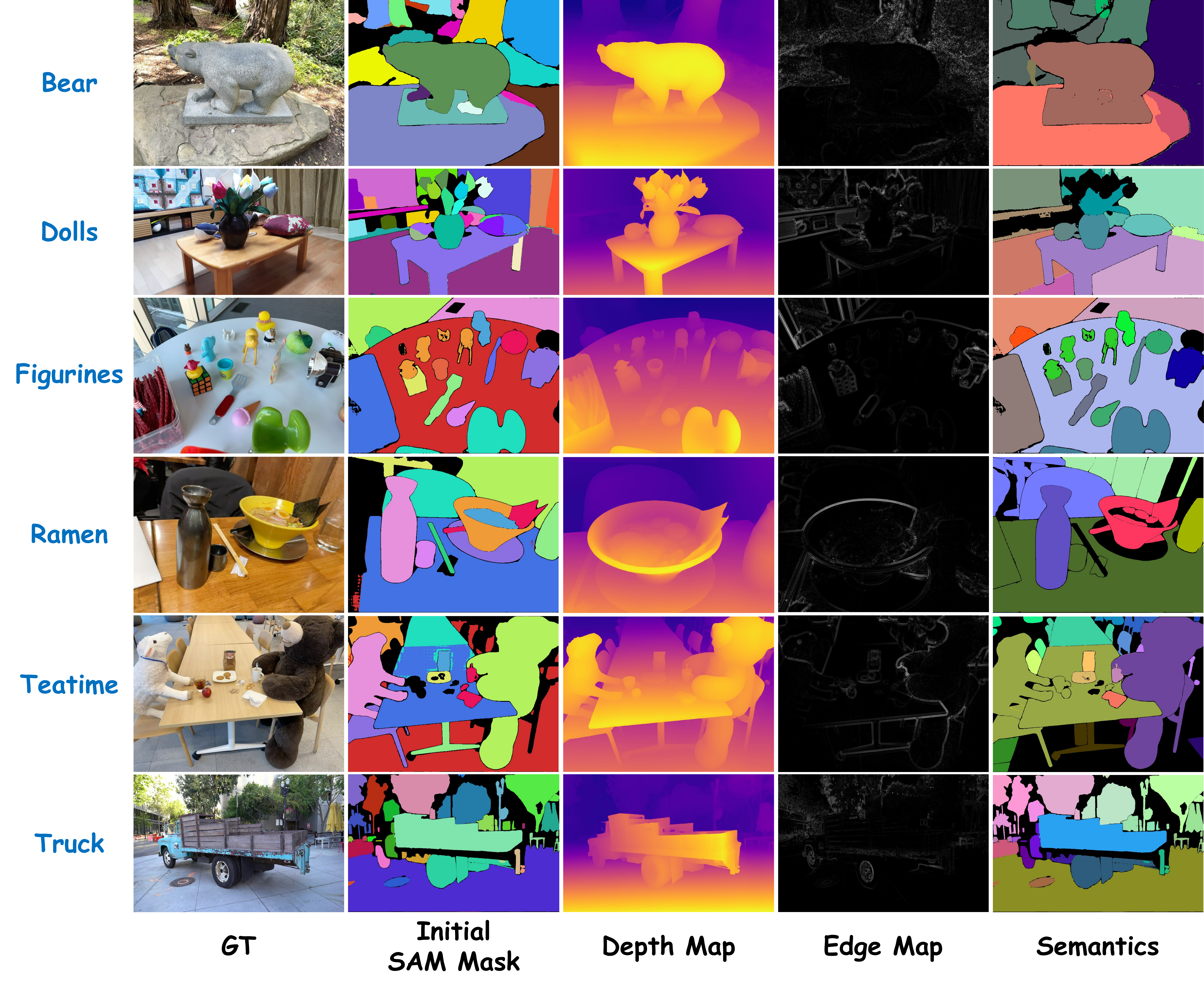}
    \caption{\textbf{Visualization of multi-cue features.} We extract complementary priors to address the limitations of the 2D-centric proposals of SAM. 
    While the Initial SAM Mask exhibits severe oversegmentation, our framework leverages Depth Maps (geometric context), Edge Maps (structural discontinuities), and Semantics (high-level object coherence) to consolidate fragmented regions. 
    This multi-cue integration ensures that the resulting segments align precisely with physical object boundaries across diverse scenes}
    \label{fig:multicue}
\end{figure}

\subsection{Semantic Feature Extraction: DINOv2}
The choice of DINOv2~\cite{dinov2} over language-aligned models such as CLIP~\cite{CLIP} or LSeg~\cite{lseg} is driven by the requirement for spatially dense representations. 
CLIP is optimized for global image-text alignment, resulting in coarse feature maps that lack the local geometric precision required for lifting features to individual 3D Gaussians. 
In contrast, DINOv2 provides patch-level features (vitb14, 768D) that preserve the local structure of scenes. 
While SAM features are excellent for boundary localization, they often lack the semantic depth required to associate fragments across multiple views. 
DINOv2’s self-supervised pre-training enables it to capture fine-grained visual similarities (e.g., different parts of a complex chair) that are often absent in text-supervised models, providing a more stable foundation for our feature lifting and consensus voting stages.

\subsection{Edge Detection: Laplacian-of-Gaussian (LoG)}
For structural boundary detection, we utilize the Laplacian of Gaussian (LoG)~\cite{LoG} instead of modern learning-based detectors or multi-stage algorithms such as Canny~\cite{Canny}. 
A key requirement for our composite score $S(i,j)$ is a continuous edge response. 
While Canny employs non-maximum suppression and hysteresis thresholding to produce binary edges, LoG provides a continuous strength map. 
This allows our scoring mechanism to weigh edge proximity probabilistically rather than being forced into a binary decision that may introduce artifacts.
Theoretically, the second-derivative formulation of the LoG is rotation invariant, making it superior to directional gradient operators (such as Sobel~\cite{Sobel}) when processing complex 3D scenes with arbitrary edge orientations. 
We utilize a Gaussian kernel with $\sigma=2.0$ to perform multi-scale filtering, which suppresses high-frequency texture noise (e.g., a patterned rug) while preserving salient structural boundaries of objects. 
This parameter-free approach ensures that our edge detection remains robust across different datasets without requiring the retraining or fine-tuning that learning-based detectors would necessitate.
\subsection{Synthesis of Multi-Cues}
The integration of these three models addresses the fundamental limitations of single-modality reasoning. 
While depth resolves appearance aliasing, semantics capture high-level coherence, and edge detection identifies boundaries that may be obscured by semantic similarities. 
Specifically, the complementary nature of these cues facilitates robust decision-making in complex environments. 
For instance, in texture-less regions where semantic embeddings ($f_i$) may appear near-identical, the monocular depth prior ($\bar{d}_i$) provides critical evidence for foreground-background separation via a hard depth constraint. 
Simultaneously, the LoG-based edge maps ($\nabla_\text{edge}$) act as structural stabilizers, preventing the overconsolidation of adjacent objects that share similar depth values but belong to distinct semantic categories.
This tripartite synergy ensures that our "actionable primitives" are both geometrically accurate and semantically consistent across all views. 
By synthesizing these signals, the framework effectively resolves the "white-on-white dilemma" and preserves the fine-grained structural integrity that single-view 2D priors typically overlook.

% ------------------------------------------------------------------------------------------------------
\section{Additional Qualitative Results}
\label{sec:additional_quali}

\begin{figure}[t]
    \centering
    \includegraphics[width=1\linewidth]{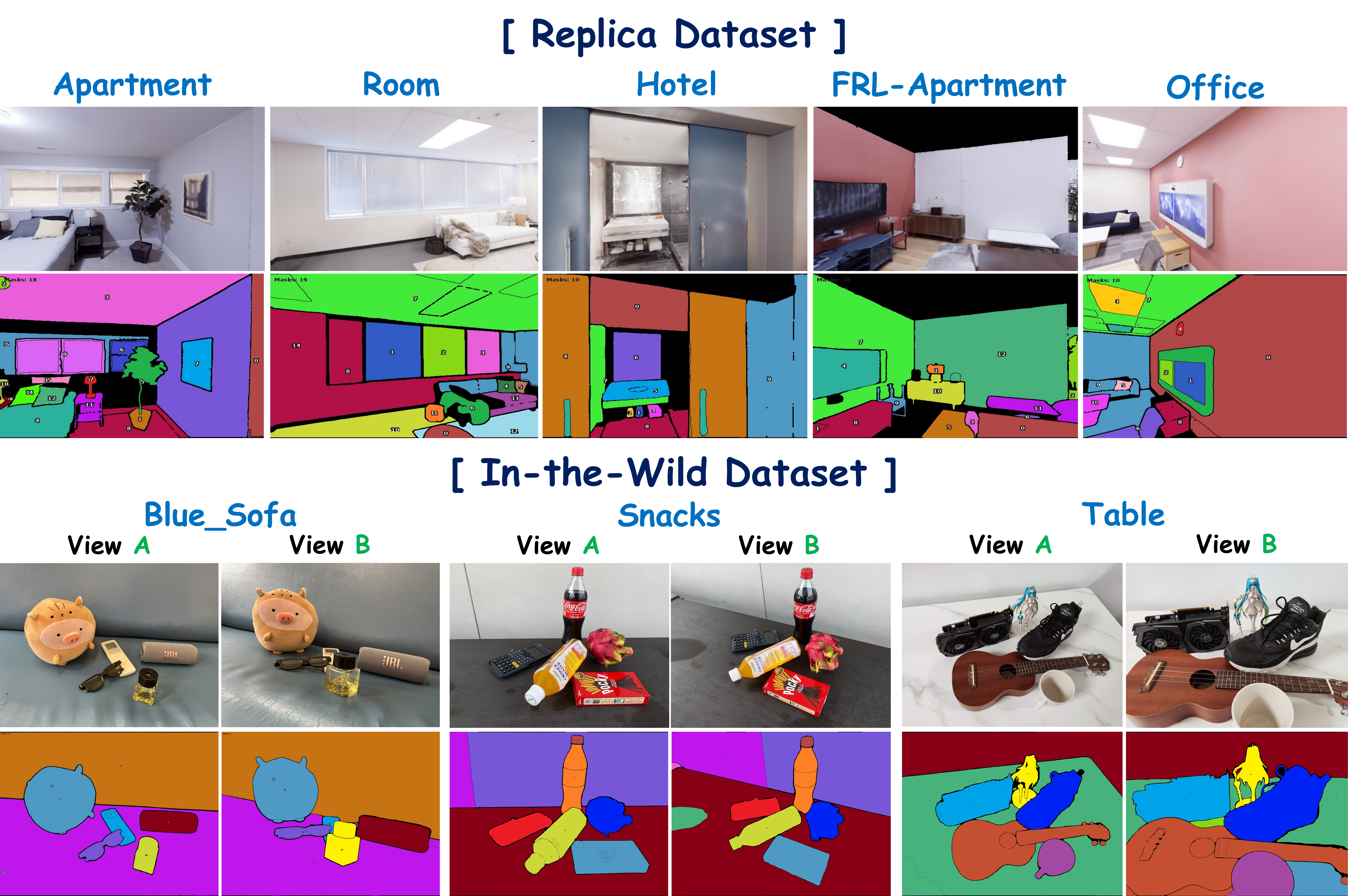}
    \caption{\textbf{Additional Qualitative Result.}
    Our method successfully generates globally coherent 3D instance masks across various indoor and real-world scenes.}
    \label{fig:additional_qual}
\end{figure}

\noindent We provide further qualitative evidence to demonstrate the robustness and generalization of our proposed framework across a variety of 3D environments, including synthetic indoor scenes and real-world captures.\\
\textbf{Generalization on the Replica Dataset.}
The upper row of our qualitative gallery showcases the performance of our method on the Replica dataset~\cite{straub2019replica}, which comprises diverse indoor environments such as Apartment, Room, Hotel, FRL-Apartment, and Office. 
As illustrated in Fig. ~\ref{fig:additional_qual}, our framework consistently produces semantically meaningful 3D instances across these varied layouts. 
Specifically, the model effectively isolates individual furniture items and architectural elements with high precision, even in cluttered office or residential settings. 
By leveraging the synergy between DINOv2-based semantic features and geometric priors, our approach successfully resolves the oversegmentation typical of 2D foundation models, ensuring that large surfaces, such as walls and floors, are consolidated into singular, coherent masks.\\
\textbf{Consistency in In-the-Wild Scenarios.}
To verify the practical applicability of our method, we evaluate it on "In-the-Wild" datasets featuring real-world objects in uncontrolled lighting and complex backgrounds, as shown in the bottom panel. 
In the Blue\_Sofa, Snacks, and Table sequences, we observe that our method maintains remarkable identity stability across significant viewpoint transitions (View A and View B). 
Unlike per-view segmentation baselines that suffer from flickering labels, our framework ensures that each object—ranging from intricate snack packaging to complex arrangements of items on a table—retains a consistent identity and sharp boundary across different camera angles. 
These results confirm that our joint reasoning over semantic, depth, and edge cues provides a robust foundation for stable object-level scene understanding in diverse real-world configurations.

\begin{table}[t]
    \centering
    \caption{\textbf{Sensitivity Analysis of Hyperparameters.} We evaluate the impact of merging thresholds ($\tau$) and feature weights ($w$) on the Replica dataset. The \textbf{Ours} configuration achieves the optimal balance between object coherence and reconstruction fidelity.}
    \label{tab:supp_hyper}
    \footnotesize
    \setlength{\tabcolsep}{3.5pt}
    \resizebox{\columnwidth}{!}{
    \begin{tabular}{lccc|ccc|cc}
        \toprule
        \multirow{2}{*}{Config.} & \multirow{2}{*}{$\tau_{merge}$} & \multirow{2}{*}{$\tau_{depth}$} & \multirow{2}{*}{$w_{sem}$} & \multicolumn{3}{c|}{Rendering Metrics} & \multicolumn{2}{c}{Segmentation} \\
        \cmidrule(lr){5-7} \cmidrule(lr){8-9}
        & & & & PSNR $\uparrow$ & SSIM $\uparrow$ & LPIPS $\downarrow$ & mIoU $\uparrow$ & \# Mask $\downarrow$ \\
        \midrule
        Aggressive Merging & 0.70 & 0.30 & 0.5 & 27.8 & 0.88 & 0.15 & 0.624 & \textbf{32} \\
        Conservative Merging & 0.95 & 0.05 & 0.5 & \textbf{28.6} & 0.90 & 0.13 & 0.691 & 412 \\
        \midrule
        Structural-biased & 0.85 & 0.15 & 0.2 & 27.9 & 0.89 & 0.14 & 0.652 & 89 \\
        Semantic-biased & 0.85 & 0.15 & 0.8 & 28.3 & 0.90 & 0.13 & 0.685 & 124 \\
        \midrule
        \textbf{Ours} & \textbf{0.85} & \textbf{0.15} & \textbf{0.5} & \textbf{28.6} & \textbf{0.91} & \textbf{0.12} & \textbf{0.728} & 67 \\
        \bottomrule
    \end{tabular}%
    }
\end{table}

\section{Ablation Studies}
\label{sec:Ablation}
Table~\ref{tab:supp_hyper} presents a comprehensive sensitivity analysis of our framework's key hyperparameters, including merging thresholds ($\tau_{merge}, \tau_{depth}$) and semantic feature weights ($w_{sem}$). 
We evaluate these configurations to identify the optimal balance between photometric fidelity and object coherence.\\
\noindent\textbf{Impact of Merging Thresholds.}
The choice of thresholds significantly dictates the level of mask consolidation. 
Aggressive Merging ($\tau_{merge}=0.70$) achieves the most compact representation with only 32 masks; however, it suffers from a substantial drop in segmentation accuracy ($0.624$ mIoU) and rendering quality ($27.8$ PSNR). 
This performance degradation stems from the overconsolidation of physically distinct objects that share similar semantic or geometric signatures. 
Conversely, Conservative Merging ($\tau_{merge}=0.95$) maintains high photometric fidelity but fails to resolve the fundamental oversegmentation of SAM, resulting in an excessive mask count of 412.\\
\noindent\textbf{Impact of Feature Weights.}
The balancing weight $w_{sem}$ modulates the influence of the DINOv2 semantic embeddings relative to the geometric and structural cues. 
A structure-biased configuration ($w_{sem}=0.2$) leads to fragmented results because it overlooks object-level semantic continuity. 
Conversely, a semantic-biased setup ($w_{sem}=0.8$) tends to ignore geometric boundaries, causing "label bleeding" across depth discontinuities.\\
\noindent\textbf{Optimal Configuration.}
Our default configuration (Ours) employs $\tau_{merge}=0.85, \tau_{depth}=0.15,$ and $w_{sem}=0.5$. 
This setup achieves a state-of-the-art balance, yielding the highest mIoU ($0.728$) and SSIM ($0.91$) while effectively consolidating the scene into $67$ actionable primitives. 
This confirms that our multi-cue scoring framework is the most robust when the semantic and geometric signals are weighted equally.

\section{Evaluation Metrics}
\label{sec:metrics}
We employ multiple metrics to evaluate segmentation quality, cross-view consistency, and editing applicability.

\noindent\textbf{Mask Intersection-over-Union (mIoU).}
The standard metric for segmentation quality is as follows: Given the predicted mask $M_p$ and ground truth mask $M_g$:
\begin{equation}
    \text{IoU}(M_p, M_g) = \frac{|M_p \cap M_g|}{|M_p \cup M_g|}.
\end{equation}
For multi-object scenes, we compute mean IoU (mIoU) by matching predicted and ground truth masks using the Hungarian algorithm~\cite{kuhn1955hungarian}:
\begin{equation}
    \text{mIoU} = \frac{1}{N} \sum_{i=1}^{N} \text{IoU}(M_p^{i}, M_g^{\pi(i)}),
\end{equation}
where $\pi$ is the optimal assignment, and $N$ is the number of ground-truth objects.

\noindent\textbf{Boundary mIoU.}
To evaluate boundary quality, we compute IoU only within a narrow band around object boundaries. 
Let $\mathcal{B}_\epsilon(M)$ denote the $\epsilon$-dilated boundary of mask $M$:
\begin{equation}
    \mathcal{B}_\epsilon(M) = (M \oplus \epsilon) \setminus (M \ominus \epsilon),
\end{equation}
where $\oplus$ and $\ominus$ denote dilation and erosion operations, respectively, with a disk kernel of radius $\epsilon = 5$ pixels. 
This metric is crucial for 3DGS-based editing, where sharp boundaries are required to prevent texture bleeding during the manipulation of objects.
Boundary mIoU is:
\begin{equation}
    \text{Boundary-mIoU} = \frac{1}{N} \sum_{i=1}^{N} \frac{|(M_p^{i} \cap M_g^{\pi(i)}) \cap \mathcal{B}_\epsilon(M_g^{\pi(i)})|}{|(M_p^{i} \cup M_g^{\pi(i)}) \cap \mathcal{B}_\epsilon(M_g^{\pi(i)})|}.
\end{equation}
\noindent\textbf{Mask Count (\#Mask).}
To quantify the degree of over-segmentation, which is a primary limitation of raw SAM proposals, we recorded the total number of unique masks $|\mathcal{M}|$ generated for each scene. 
A lower \#Mask value, when paired with a high mIoU, indicates that the framework successfully consolidates fragmented regions into coherent, semantically meaningful object instances.

\section{Limitations and Future Works}
\label{sec:LimitFuture}
\subsection{Limitations} 
Despite its improved coherence, our framework has two primary limitations. 
First, it assumes that SAM~\cite{SAM} provides an oversegmented initial state. 
Thus, severe undersegmentation or missed boundaries are irrecoverable through merging alone. 
Second, because per-view merging operates independently before global clustering, extreme occlusions or appearance changes may yield localized inconsistencies.

\subsection{Future Works} 
Our pipeline significantly enhances index-based editing in frameworks such as GaussianGrouping~\cite{gaussiangrouping} by providing unified "\textit{actionable primitives}" and eliminating manual fragment selection. 
In the future, we aim to utilize these consistent 3D masks as robust conditions for generative tasks, such as high-fidelity object replacement, which currently suffer from boundary artifacts. 
Furthermore, extending our multi-cue reasoning to dynamic 3DGS could enable temporally stable tracking and semantic manipulation in complex spatiotemporal scenes. 
Finally, integrating language-guided models, such as CLIP~\cite{CLIP} or LangSplat~\cite{langsplat} with our refined feature fields will allow for intuitive text-driven object disambiguation, further bridging the gap between raw scene representation and user-centric 3D content creation.

% 이부분은 ICPR paper submission guideline에 LLM tool 사용 여부를 기재하라고 되어 있어서 그림 작업 때 사용했다는 점을 기재했습니다
\begin{figure}[t]
    \centering
    \includegraphics[width=1\linewidth]{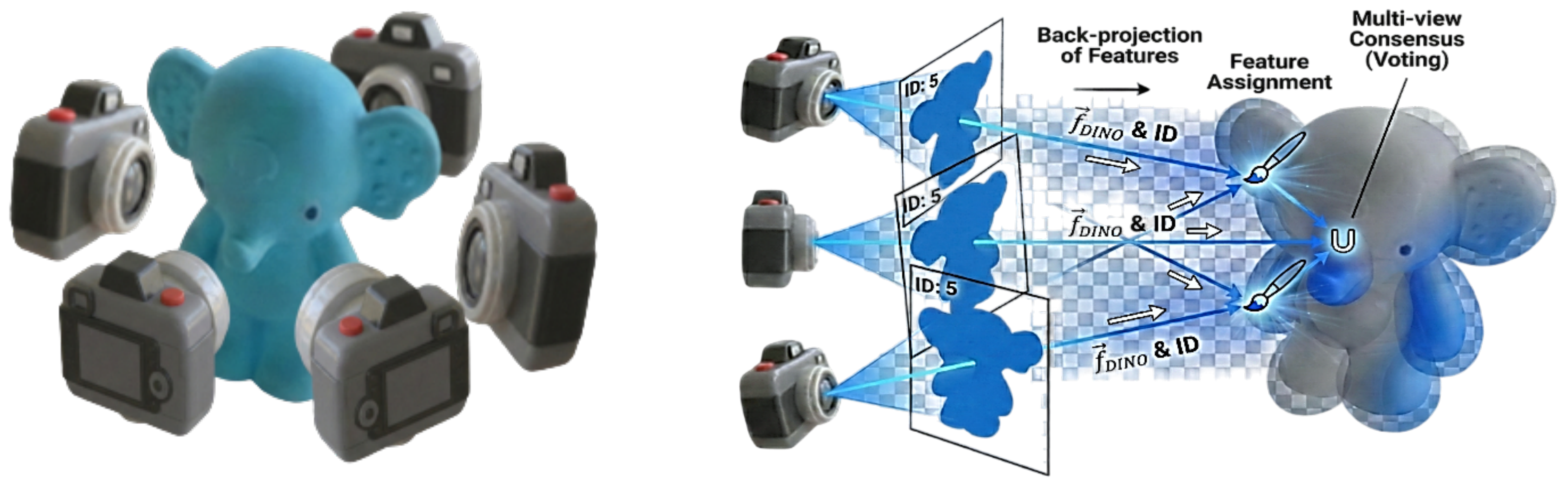}
    \caption{\textbf{AI generated Images.}
    We utilized these images in our main pipeline to illustrate effectively.
    }
    \label{fig:aigen}
\end{figure}

\section{Generative AI Disclosure} 
The authors utilized the Gemini Pro model to assist in generating the illustrative components of the pipeline diagram, specifically for visualizing the camera perspectives that surround the target object (elephant) and the back-projection of features into 3D Gaussians ($G=\{\mu,\Sigma,\alpha,C\}$), as shown in Fig.~\ref{fig:aigen}. 
To ensure transparency, we provide the specific prompts used for the generation process:
\begin{itemize}
    \item \textbf{Prompt 1 (Object Isolation):} \\"An illustration showing a target object (an elephant) isolated from a complex scene using a segmentation mask, with multiple camera frustums from different angles pointing at the object to represent a multiview capture setup."
    \item \textbf{Prompt 2 (Feature Lifting):} \\"A conceptual diagram showing the back-projection of 2D object features into a 3D space. 
    The 2D masks with global IDs are lifted and integrated into 3D Gaussian primitives, illustrating the formation of a view-consistent 3D feature field."
\end{itemize}
The AI-generated components were used solely for the visual communication of technical concepts and did not contribute to the data analysis or the scientific results of this study.

% ==================================================================================================

\section*{Acknowledgements}
This work was partly supported by Institute of Information \& communications Technology Planning \& Evaluation (IITP) grant funded by the Korea government(MSIT) (No.RS-2025-25422680, Metacognitive AGI Framework and its Applications, 33 \%), Institute of Information \& communications Technology Planning \& Evaluation (IITP) grant funded by the Korea government(MSIT) (No.RS-2020-II201373, Artificial Intelligence Graduate School Program(Hanyang University), 33 \%) and the National Research Foundation of Korea(NRF) grant funded by the Korea government(MSIT) (RS-2025-00521432, 34 \%).

{
    \small
    \bibliographystyle{splncs04}
    \bibliography{main}
}

\end{document}